\documentclass{article}
\usepackage[utf8]{inputenc}
\usepackage[margin=1in]{geometry}

\usepackage{times}
\usepackage{parskip}
\usepackage{authblk}

\usepackage{amssymb}
\usepackage{amsmath}
\usepackage{tikz-cd}
\usepackage{mathtools}

\usepackage{algorithm}
\usepackage[noend]{algpseudocode}

\usepackage[colorlinks=true,urlcolor=blue]{hyperref}

\usepackage{natbib}
\usepackage{usebib}
\usepackage{cancel}
\bibinput{references}

\usepackage[table, dvipsnames]{colortbl}
\usepackage{multirow}
\usepackage{booktabs}
\usepackage{subcaption}

\title{BabyAI 1.1}
\author[1]{David~Yu-Tung~Hui}
\author[1]{Maxime~Chevalier-Boisvert}
\author[2]{Dzmitry~Bahdanau}
\author[1, 3]{Yoshua~Bengio}
\affil[1]{Mila, Universit\'e de Montr\'eal}
\affil[2]{Element AI}
\affil[3]{CIFAR Senior Fellow}
\date{\today}

\begin{document}
\setcitestyle{square}
\maketitle

\begin{abstract}
    The BabyAI platform is designed to measure the sample efficiency of training an agent to follow grounded-language instructions.  BabyAI 1.0 presents baseline results of an agent trained by deep imitation or reinforcement learning.  BabyAI 1.1 improves the agent's architecture in three minor ways.  This increases reinforcement learning sample efficiency by up to $3 \times$ and improves imitation learning performance on the hardest level from $77 \%$ to $90.4 \%$.  We hope that these improvements increase the computational efficiency of BabyAI experiments and help users design better agents.
\end{abstract}

\section{Introduction}

The BabyAI platform \footnote{(\url{https://github.com/mila-iqia/babyai}} is an environment designed to evaluate how well an agent follows grounded-language instructions.  The quality of an agent is measured with two metrics: its success rate at following instructions and the number of episodes or demonstrations required to train it.  BabyAI 1.0, \citep{babyai_iclr19}, presents results of a baseline agent trained by reinforcement and imitation learning (RL and IL) methods.  In this technical report, we present three modifications that significantly improved the baseline results.

Two modifications are to the network's architecture and the third to the representation of the visual input.  The network is modified by removing maxpooling at lower levels in the visual encoder and adding residual connections around FiLM layers \citep{perez_film:_2017}.  The visual representation is modified to use learned embeddings in a Bag-of-Words fashion \citep{mikolov2013efficient}.

\section{Proposed Architectural and Representational Modifications}
\label{sec:arch}

This section describes the network architecture and BabyAI 1.0 visual representation before detailing the two architectural modifications and two alternate visual representations.
\begin{figure}[!ht]
    \centering
    \def\svgwidth{.8\columnwidth}
    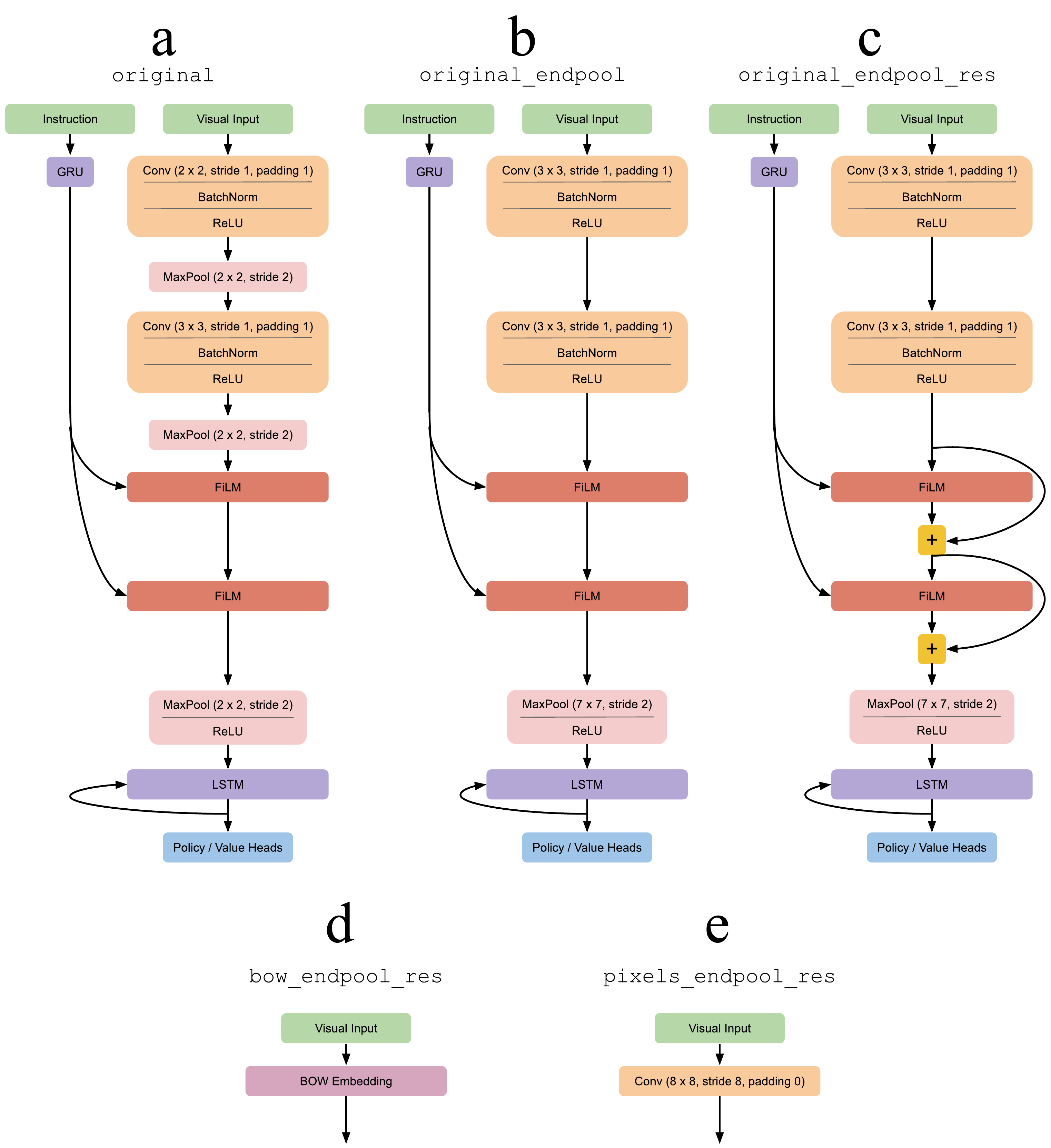
    \caption{The five architectures. (a), (b) and (c): BabyAI 1.0 architecture, first modification removing pooling in the visual encoder and second modification introducing residual connections around FiLM. (d), (e): respective architectural changes needed to use BOW and pixel visual representations.  Arrows at the bottom of architectures (d) and (e) feed into an architecture identical from the first convolutional layer of (c) onwards.}
    \label{fig:models}
\end{figure}
The BabyAI platform has nineteen levels which can be categorised into two types: small and big \citep{babyai_iclr19}.  Small levels are single-room but big levels are usually $3 \times 3$ rooms.  The BabyAI 1.0 baseline agent has two architectures used on the small and big levels.  These architectures have the same structure and are illustrated by Figure \ref{fig:models}.a.  The architecture takes two inputs, a visual input and a linguistic instruction.  We use FiLM to combine the outputs of a convolutional `visual encoder' with a GRU \citep{cho_learning_2014} embedding of the instruction.  We refer the reader to Appendix \ref{section:arch} for more details concerning the distinction between `big' and `small'. 

Figures \ref{fig:models}.b and \ref{fig:models}.c respectively present two architectural modifications: removing pooling in the visual encoder and adding residual connections around the image convolution and the FiLM layers.  To ensure that the shape of the visual encoder is consistent after pooling is removed, we change filter size from $2\times2$ to $3\times3$.  We expect these changes to improve sample efficiency because they enable more information to be transmitted to higher layers.

At every timestep, the agent receives visual information about a $7 \times 7$ grid of tiles which are immediately in the direction it is facing.  BabyAI 1.0 represents a tile by a triple-integer value.  The first integer describes the type of object in the tile and the second integer the object's color.  The third integer is only used if the object is a door, and describes whether it is open, closed or locked.  BabyAI 1.0 represents a visual input by concatenating all tile representations together.  This results in a tensor of size $7 \times 7 \times 3$.

A gridworld tile (and thus the visual input) can also be represented in two other ways: as a ``bag of words'' or by an RGB image.
In the Bag-of-Words (``BOW'') approach a set of symbols that describes the tile is embedded in a trainable lookup table. This approach is commonly used in gridwords such as \citep{leike2017ai}, \citep{rajendran2015attend} and \citep{SchraderSokoban2018}.  Because a tile in BabyAI can be represented by three integers, we use three look-up tables and use each integer as a key.  A tile is then represented by the mean of the three looked-up feature vectors.  As with the BabyAI 1.0 visual representation, the BOW representation is formed from combining all tile representations into a 3D tensor.  As we set the dimensionality of a feature vector to 128, the dimensionality of the BOW visual representation is $7 \times 7 \times 128$.  This is depicted in Figure \ref{fig:models}.d.

The contents of a tile can also be represented by a 3-channel Red-Green-Blue (RGB) image.  As we choose the size of an image to be $8$ pixels, a grid is thus represented by an image stored in a $8 \times 8 \times 3$ tensor.  The entire $7 \times 7$ visual input can be represented in a 3-channel RGB image with dimensionality $56 \times 56 \times 3$.  An architecture using this visual representation is illustrated in Figure \ref{fig:models}. 

Architectures names are structured in two parts.  The first part is either ``original'', ``bow'' or ``pixels'', indicating which visual representation is used.  The second part is optional and describes whether ``\_endpool'' (because the only source of pooling is at the end) or ``\_res'' (adding residual connections) are present in the architecture.

\section{Experiments}

To determine the best architecture and visual representation, we follow BabyAI 1.0 and experiment on the six easiest BabyAI levels.  These six levels consist of five single-room levels and one multi-room level.  Then, we present IL performance benchmarks on all levels.

\subsection{Finding the Best Architecture}

We measure RL sample and computational efficiency and IL performance with varying number of demonstrations.  RL experiments were structured in two stages.

The first set of experiments investigated architectural modifications.  Results in Table~\ref{tbl:arch} showed that removing pooling in the visual encoder had a significant improvement on sample efficiency, but adding residual connections effected both increases and decreases.  Nevertheless, we adopted residual connections for further experiments because the sample efficiency increase for PutNextLocal greatly outweighted the total decrease in GoToLocal and GoTo.

The second set of experiments investigated visual representations.  Results in Table \ref{tbl:visual} still do not show a wide variation in sample efficiency.  Because training from pixels was hard on the two most difficult levels (GoTo, PutNextLocal), we halved the learning rate ($\alpha$ in Adam \citep{kingma_adam:_2015} from $1 \times 10^{-4}$ to $5 \times 10^{-5}$) and reran the second set of experiments.  The resulting statistics in Table \ref{tbl:visual2} do not show much variation between the three visual representations.

We now consider the computational efficiency of training each of these five architectures.  Training from pixels has a slower throughput than the other visual representations (Table \ref{tbl:fps}).  Because of this and no clear advantage in RL sample efficiency (Tables \ref{tbl:visual}, \ref{tbl:visual2}), we drop further experiments on pixels.

Now, we investigate whether changing the visual representation to BOW and two architectural modifications improve IL performance.  \citet{babyai_iclr19} measures sample efficiency using an interpolated function fitted with a Gaussian Process (GP) \citep{rasmussen_gaussian_2005}.  In our experiments we found that an infeasibly large number of training runs would be required in order to obtain a sufficiently confident sample efficiency estimate from the GP.  Instead, we follow Table 6 in \citep{zolna2020combating}, who evaluate IL by observing its success rate trained with varying number of demonstrations.  \citet{zolna2020combating} use $1/64$\textsuperscript{th}, $1/8$\textsuperscript{th} and all of 1 million demonstrations.  We use 5, 10, 50, 100 and 500 thousand demonstrations, which correspond to $1/200$\textsuperscript{th}, $1/100$\textsuperscript{th}, $1/20$\textsuperscript{th}, $1/10$\textsuperscript{th} and $1/2$\textsuperscript{th} of the total 1 million demonstrations.

IL results in Table~\ref{tbl:il} show that training from BOW is advantageous to the original BabyAI 1.0 visual representation.  Interestingly, we find that for hard levels with a few number of demos, the architectural modifications are not beneficial for training.  This is offset by changing the visual representation to BOW.

\subsection{Benchmarking the Best Modifications}

Having constructed the BabyAI 1.1 agent, we benchmark its performance over all nineteen BabyAI levels.

Table \ref{tbl:baseline} shows that modifications found by the previous section yielded improvements in performance over all levels.  Four more levels (Unlock, Putnext, Synth and SynthLoc) were solved, and success rate increased by $13.7 \%$ in the hardest level (BossLevel) from $77 \%$ to $90.4 \%$.

\section{Conclusion}

As BabyAI was intended to be a lightweight experimental platform, BabyAI 1.0 used a specific hand-crafted representation rather than a more realistic pixel-based representation.  We have shown that training from other visual representations (BOW and pixels) is feasible, and is sometimes more sample efficient (Table \ref{tbl:il}).  However, learning from pixels took longer to compute (Table \ref{tbl:fps}) and was more sensitive to hyperparameters. Besides, using pixel representations for the tiles still does not bridge the reality gap between the gridworld and the 3D real world, for almost all challenging aspects of visual perception, such as e.g. occlusion, illumination, different viewpoints are still not modelled. For this reason we keep the BabyAI input representation symbolic, though we switch to the more standard BOW approach for encoding the symbolic input.

This report introduces BabyAI 1.1, the latest version of the BabyAI platform.  This modifies the previous version of BabyAI with minor changes to the baseline agent, but major improvements to baseline statistics.  We hope that this change encourages researchers to (re-)use similar architectures within novel agents, so that research into grounded language learning may be conducted in more computationally efficient ways.

\subsection*{Acknowledgements}

This research was mostly performed at Mila with funding by the Government of Quebec and CIFAR, and enabled by Compute Canada (\url{www.computecanada.ca}).

\begin{table}
\caption{RL sample efficiency (mean $\pm$ std, thousands of episodes) in the six easiest BabyAI levels with respect to different \textit{architectures} (Table \ref{tbl:arch}) and \textit{visual representations} with 1e-4 learning rate (Table \ref{tbl:visual}) and 5e-5 learning rate (Table \ref{tbl:visual2}).  Each figure is an average over ten training runs initialised to a randomly chosen seed.  All experiments were run with the `small' architecture.  More experimental details are in Appendix \ref{section:rl_exps}.}

\begin{subtable}{\textwidth}
    \centering
    \caption*{A cell is shaded depending on whether its values are statistically significant from those of the cell \textit{to its left}.  Statistical significance is computed using a two-tailed T-test with inequal variance.}
    \begin{tabular}{ll}
        & accept null hypothesis that there is \textit{no significant difference} \\
        \cellcolor{cyan} & reject null hypothesis at 1 \% significance due to \textit{significant increase in sample efficiency} (number is smaller) \\
        \cellcolor{yellow} & reject null hypothesis at 1 \% significance due to \textit{significant decrease in sample efficiency} (number is bigger) \\
    \end{tabular}
\end{subtable}
\bigskip

\begin{subtable}{\textwidth}
    \centering
    \caption{RL sample efficiency for different architectures.  See the last paragraph of Section \ref{sec:arch} for the explanation of architecture names.}
    \label{tbl:arch}
    \begin{tabular}{lccc} 
        \toprule
        Level & \multicolumn{3}{c}{Sample Efficiency} \\
        \cmidrule(lr){2-4}
        & original & original\_endpool & original\_endpool\_res \\ 
        \midrule
        GoToRedBallGrey & 21 $\pm$ 5 & 21 $\pm$ 6 & 21 $\pm$ 5 \\
        GoToRedBall & 273 $\pm$ 27 & \cellcolor{cyan} 200 $\pm$ 16 & 179 $\pm$ 17 \\
        GoToLocal & 1311 $\pm$ 251 & \cellcolor{cyan} 381 $\pm$ 30 & \cellcolor{yellow} 437 $\pm$ 45 \\
        PickupLoc & 1797 $\pm$ 290 & \cellcolor{cyan} 743 $\pm$ 132 & 710 $\pm$ 166 \\
        PutNextLocal & 2984 $\pm$ 172 & \cellcolor{cyan} 2169 $\pm$ 739 & \cellcolor{cyan} 1009 $\pm$ 128 \\
        GoTo & 1601 $\pm$ 463 & \cellcolor{cyan} 454 $\pm$ 69 & \cellcolor{yellow} 813 $\pm$ 278 \\
        \bottomrule
    \end{tabular}
\end{subtable}
\bigskip

\begin{subtable}{\textwidth}
    \centering
    \caption{RL sample efficiency for different visual representations with \textit{learning rate = $10^{-4}$}.  See the last paragraph of Section \ref{sec:arch} for the explanation of architecture names.}
    \label{tbl:visual}
    \vspace{0.1cm}
    \begin{tabular}{lccc}
        \toprule
        Level & \multicolumn{3}{c}{Sample Efficiency} \\
        \cmidrule(lr){2-4}
        & original\_endpool\_res & bow\_endpool\_res & pixels\_endpool\_res \\
        \midrule
        GoToRedBallGrey & 21 $\pm$ 5 & 24 $\pm$ 2 & \cellcolor{yellow} 34 $\pm$ 4 \\
        GoToRedBall & 179 $\pm$ 17 & 177 $\pm$ 2 & \cellcolor{cyan} 172 $\pm$ 2 \\
        GoToLocal & 437 $\pm$ 45 & 611 $\pm$ 760 & 242 $\pm$ 15 \\
        PickupLoc & 710 $\pm$ 166 & 982 $\pm$ 266 & 1082 $\pm$ 385 \\
        PutNextLocal & 1092 $\pm$ 143 & \cellcolor{cyan} 876 $\pm$ 104 & Not Trainable \\
        GoTo & 813 $\pm$ 278 & 817 $\pm$ 502 & Not Trainable \\
        \bottomrule
    \end{tabular}
\end{subtable}
\bigskip

\begin{subtable}{\textwidth}
    \centering
    \caption{RL sample efficiency for different visual representations with \textit{learning rate = $5 \times 10^{-5}$}.  See the last paragraph of Section \ref{sec:arch} for the explanation of architecture names.}
    \label{tbl:visual2}
    \begin{tabular}{lccc}
        \toprule
        Level & \multicolumn{3}{c}{Sample Efficiency} \\
        \cmidrule(lr){2-4}
        & original\_endpool\_res & bow\_endpool\_res & pixels\_endpool\_res \\
        \midrule
        GoToRedBallGrey & 35 $\pm$ 5 & 30 $\pm$ 4 & \cellcolor{yellow} 44 $\pm$ 11 \\
        GoToRedBall & 263 $\pm$ 22 & \cellcolor{cyan} 164 $\pm$ 3 & \cellcolor{cyan} 155 $\pm$ 5 \\
        GoToLocal & 606 $\pm$ 81 & 449 $\pm$ 176 & 336 $\pm$ 28 \\
        PickupLoc & 1732 $\pm$ 579 & 1461 $\pm$ 422 & 1308 $\pm$ 421 \\
        PutNextLocal & 1277 $\pm$ 252 & 876 $\pm$ 104 & 1301 $\pm$ 320 \\
        GoTo & 984 $\pm$ 484 & 803 $\pm$ 525 & 845 $\pm$ 329 \\
        \bottomrule
    \end{tabular}
\end{subtable}
\end{table}

\begin{table}
    \centering
    \caption{Frames Per Second (mean $\pm$ std) of RL training with different architectures, averaged across the six easiest BabyAI levels.  Inter-level differences are negligible.}
    \label{tbl:fps}
    \vspace{0.1cm}
    \begin{tabular}{lccccc}
        \toprule
        Architecture & original & original\_endpool & original\_endpool\_res & bow\_endpool\_res & pixels\_endpool\_res \\
        \midrule
        RL (FPS) & 1139 $\pm$ 128 & 927 $\pm$ 72 & 907 $\pm$ 69 & 855 $\pm$ 58 & 540 $\pm$ 67 \\
        \bottomrule
    \end{tabular}
\end{table}

\begin{table}
\caption{IL success rate (\%) in the six easiest BabyAI levels with respect to varying number of demonstrations, architectures and visual representations.  Experiments have a success rate $\geqslant 99 \%$ are successful and are \textbf{bolded}.  Each figure is an average over ten training runs initialised to a randomly chosen seed.  Standard deviations of $\pm 0.0$ are omitted for clarity.  The first five experiments were run with the `small' architecture.  GoTo experiments were run with the `large' architecture.  More experimental details are given in Appendix~\ref{section:il_exps}.}
\label{tbl:il}

\begin{subtable}{\textwidth}
    \centering
    \caption*{A cell is shaded depending on whether its values are statistically significant from those of the cell \textit{to its left}.  Statistical significance is computed using a two-tailed T-test with inequal variance.}
    \begin{tabular}{ll}
    & accept null hypothesis that there is \textit{no significant difference} \\
    \cellcolor{cyan} & reject null hypothesis at 1 \% significance due to \textit{significant increase in performance} (number is bigger) \\
    \cellcolor{yellow} & reject null hypothesis at 1 \% significance due to \textit{significant decrease in performance} (number is smaller) \\
    \end{tabular}
\end{subtable}
\bigskip

\begin{subtable}{\textwidth}
    \centering
    \begin{tabular}{lcccc}
        \toprule
        Level & Number of Demos (thousands)  & \multicolumn{3}{c}{Success Rate} \\
        \cmidrule(lr){3-5}
        & & original & original\_endpool\_res & bow\_endpool\_res \\
        \midrule
        GoToRedBallGrey
        & 5 & {\bf 99.5 $\pm$ 0.1} & {\bf 99.5 $\pm$ 0.1} & \cellcolor{cyan} {\bf 99.7 $\pm$ 0.1} \\
        & 10 & {\bf 99.7} & {\bf 99.8 $\pm$ 0.1} & \cellcolor{cyan} {\bf 99.9} \\
        & 50 & {\bf 100} & \cellcolor{cyan} {\bf 100} & {\bf 100} \\
        & 100 & {\bf 100} & \cellcolor{cyan} {\bf 100} & \cellcolor{cyan} {\bf 100} \\
        & 500 & {\bf 100} & {\bf 100} & {\bf 100} \\
        \midrule
        GoToRedBall
        & 5 & 89.6 $\pm$ 0.3 & \cellcolor{cyan} 91.3 $\pm$ 0.6 & \cellcolor{cyan} {\bf 99.3 $\pm$ 0.3} \\
        & 10 & 93.1 $\pm$ 0.8 & \cellcolor{cyan} 95.6 $\pm$ 0.7 & \cellcolor{cyan} {\bf 99.8 $\pm$ 0.1} \\
        & 50 & {\bf 99.2 $\pm$ 0.2} & \cellcolor{cyan} {\bf 99.9} & \cellcolor{cyan} {\bf 100} \\
        & 100 & {\bf 99.7} & \cellcolor{cyan} {\bf 100} & {\bf 100} \\
        & 500 & {\bf 99.9} & \cellcolor{cyan} {\bf 100} & {\bf 100} \\
        \midrule
        GoToLocal
        & 5 & 72.5 $\pm$ 1.0 & 71.6 $\pm$ 1.4 & \cellcolor{cyan} 84.2 $\pm$ 2.0 \\
        & 10 & 79.9 $\pm$ 1.2 & 79.7 $\pm$ 1.8 & \cellcolor{cyan} 94.2 $\pm$ 0.8 \\
        & 50 & 95.3 $\pm$ 0.5 & \cellcolor{cyan} {\bf 99.6 $\pm$ 0.1} & \cellcolor{cyan} {\bf 99.8 $\pm$ 0.1} \\
        & 100 & {\bf 97.8 $\pm$ 0.3} & \cellcolor{cyan} {\bf 99.9} & {\bf 99.9} \\
        & 500 & {\bf 99.6 $\pm$ 0.1} & \cellcolor{cyan} {\bf 100} & {\bf 100} \\
        \midrule
        PutNextLocal
        & 5 & 22.3 $\pm$ 1.7 & \cellcolor{yellow} 12.0 $\pm$ 1.8 & 12.5 $\pm$ 1.2 \\
        & 10 & 39.1 $\pm$ 3.5 & \cellcolor{yellow} 16.2 $\pm$ 2.3 & \cellcolor{cyan} 24.9 $\pm$ 3.2 \\
        & 50 & 80.8 $\pm$ 1.4 & \cellcolor{cyan} 90.6 $\pm$ 3.5 & 88.6 $\pm$ 11.0 \\
        & 100 & 93.9 $\pm$ 0.4 & \cellcolor{cyan} {\bf 99.5 $\pm$ 0.1} & {\bf 99.5 $\pm$ 0.5} \\
        & 500 & {\bf 99.3 $\pm$ 0.2} & \cellcolor{cyan} {\bf 100} & \cellcolor{cyan} {\bf 100} \\
        \midrule
        PickupLoc
        & 5 & 53.0 $\pm$ 1.3 & \cellcolor{yellow} 35.9 $\pm$ 1.5 & \cellcolor{cyan} 60.3 $\pm$ 1.8 \\
        & 10 & 65.3 $\pm$ 1.5 & \cellcolor{yellow} 53.7 $\pm$ 1.2 & \cellcolor{cyan} 74.9 $\pm$ 3.9 \\
        & 50 & 90.8 $\pm$ 1.5 & \cellcolor{cyan} 96.2 $\pm$ 0.5 & \cellcolor{cyan} 97.0 $\pm$ 0.3 \\
        & 100 & 96.4 $\pm$ 0.5 & \cellcolor{cyan} {\bf 98.5 $\pm$ 0.4} & {\bf 98.6 $\pm$ 0.3} \\
        & 500 & {\bf 99.5 $\pm$ 0.2} & \cellcolor{cyan} {\bf 99.8 $\pm$ 0.1} & {\bf 99.8 $\pm$ 0.1} \\
        \midrule
        GoTo & 10 & 70.4 $\pm$ 1.1 & \cellcolor{cyan} 76.3 $\pm$ 5.0 & \cellcolor{cyan} 96.1 $\pm$ 0.4 \\ 
        & 100 & 94.9 $\pm$ 0.3 & \cellcolor{cyan} {\bf 99.3 $\pm$ 0.1} & {\bf 99.4} \\
        \bottomrule
    \end{tabular}
\end{subtable}
\end{table}


\begin{table}
    \centering
    \caption{Comparision of baseline IL results for all BabyAI levels.  Experiments with a success rate $\geqslant 99 \%$ are successful and are \textbf{bolded}.  On all levels, the `big' configuration was trained on 1 million demonstrations until the loss has converged.  As running these experiments are computationally expensive, we present results on 1 seed.  Success rate is calculated as with 512 trials once the loss has converged.}
    \label{tbl:baseline}
    \begin{tabular}{lccc}
        \toprule
        Level & \multicolumn{2}{c}{Success Rate (\%)} & Demo Length (Mean $\pm$ Std) \\
        \cmidrule(lr){2-3}
        & BabyAI 1.0 & BabyAI 1.1 & \\
        \midrule
        GoToObj & {\bf 100} & {\bf 100} & 5.18 $\pm$ 2.38 \\
        GoToRedBallGrey & {\bf 100} & {\bf 100} & 5.81 $\pm$ 3.29 \\
        GoToRedBall & {\bf 100} & {\bf 100} & 5.38 $\pm$ 3.13 \\
        GoToLocal & {\bf 99.8} & {\bf 100} & 5.04 $\pm$ 2.76 \\
        PutNextLocal & {\bf 99.2} & {\bf 100} & 12.4 $\pm$ 4.54 \\
        PickupLoc & {\bf 99.4} & {\bf 100} & 6.13 $\pm$ 2.97 \\
        GoToObjMaze & {\bf 99.9} & {\bf 100} & 70.8 $\pm$ 48.9 \\
        GoTo & {\bf 99.4} & {\bf 100} & 56.8 $\pm$ 46.7 \\
        Pickup & {\bf 99} & {\bf 100} & 57.8 $\pm$ 46.7 \\
        UnblockPickup & {\bf 99} & {\bf 100} & 57.2 $\pm$ 50 \\
        Open & {\bf 100} & {\bf 100} & 31.5 $\pm$ 30.5 \\
        Unlock & 98.4 & {\bf 100} & 81.6 $\pm$ 61.1 \\
        PutNext & 98.8 & {\bf 99.6} & 89.9 $\pm$ 49.6 \\
        Synth & 97.3 & {\bf 100} & 50.4 $\pm$ 49.3 \\
        SynthLoc & 97.9 & {\bf 100} & 47.9 $\pm$ 47.9 \\
        GoToSeq & 95.4 & 96.7 & 72.7 $\pm$ 52.2 \\
        SynthSeq & 87.7 & 93.9 & 81.8 $\pm$ 61.3 \\
        GoToImpUnlock & 87.2 & 84.0 & 110 $\pm$ 81.9 \\
        BossLevel & 77 & 90.4 & 84.3 $\pm$ 64.5 \\
        \bottomrule
    \end{tabular}
\end{table}

\pagebreak
\bibliography{references}
\bibliographystyle{apalike}
\appendix

\section{Agent Architecture}
\label{section:arch}

At every timestep, an agent receives a visual input and linguistic command of variable length which compels the agent to execute an action.  The BabyAI baseline agent is implemented by a deep neural network which processes the visual input and linguistic command, producing an action.  The visual input and linguistic instruction are respectively encoded by a Gated Recurrent Unit (GRU) and convolutional network.  These two encodings are then combined by two batch-normalised FiLM layers.  A Long-Short-Term-Memory cell (LSTM) \citep{hochreiter_long_1997} integrates the output of FiLM across timesteps.  Finally, the integrated output is passed to policy and value heads.  The agent can be trained by RL or IL methods in conjunction with BackPropagation Through Time (BPTT) \citep{werbos_bptt}.

The `small' configuration uses a unidirectional GRU and LSTM of dimensionality 128 for memory.  The `big' configuration uses a 128-dimensional bidirectional GRU with attention \citep{bahdanau_neural_2015} and the memory LSTM with dimensionality 2048. 

\section{Reinforcement Learning Experiments}
\label{section:rl_exps}

Sample efficiency is defined by the number of RL training episodes needed to train an agent to $\geqslant 99 \%$ success rate.  \citet{babyai_iclr19} defines success as whether an agent can follow an instruction within $n_{max}$ steps, a figure pre-defined for each level.

We use Advantage-Actor Critic (A2C) \citep{wu_a2c} with Proximal Policy Optimisation (PPO) \citep{schulman_proximal_2017} and Generalised Advantage Estimation (GAE) \citep{schulman_high-dimensional_2015}.  Data for A2C is collected in batches of 64 rollouts of length 40.  These were used in 4 epochs of PPO.  $\lambda=0.99$ was used in GAE.  If an agent completes a task after $n$ steps, it is rewarded with $1 - 0.9 n / n_{max}$.  Otherwise, no reward is given.  The returns were discounted by $\gamma=0.99$.  Results in Table \ref{tbl:arch} and Table \ref{tbl:visual} was optimised by Adam with the hyperparameters $\alpha=10^{-4}$, $\beta_1=0.9$, $\beta_2=0.999$ and $\epsilon=10^{-5}$.  Results in Table \ref{tbl:visual2} used $\alpha=5 \times 10^{-5}$.

\section{Imitation Learning Experiments}
\label{section:il_exps}

Different hyperparameters were used for training `small' and `big' models.

The small model was trained with a batch size of 256 and an epoch consisting of 25600 demos.  Backpropagation Through Time (BPTT) was truncated at 20 steps.

The large model had a batch size of 128 and an epoch of 102400 demonstrations.  BPTT was truncated at 80 steps.  In addition, the model was trained with an entropy regulariser, which had a coefficient of $0.01$.

These were optimised by Adam with $\alpha=10^{-4}$ for small architectures, $\alpha=5 \times 10^{-5}$ for large architectures and $\beta_1=0.9$, $\beta_2=0.999$ and $\epsilon=10^{-5}$.

\end{document}